\documentclass[conference]{IEEEtran}
\IEEEoverridecommandlockouts
\usepackage{cite}
\usepackage{amsmath,amssymb,amsfonts}
\usepackage{algorithmic}
\usepackage{graphicx}
\usepackage{textcomp}
\usepackage{xcolor}
\def\BibTeX{{\rm B\kern-.05em{\sc i\kern-.025em b}\kern-.08em
    T\kern-.1667em\lower.7ex\hbox{E}\kern-.125emX}}
\begin{document}

\title{Making sense of electrical vehicle discussions using
sentiment analysis on closely related news and user
comments\\
{\footnotesize \textsuperscript{}}
\thanks{Identify applicable funding agency here. If none, delete this.}
}

\author{\IEEEauthorblockN{1\textsuperscript{st} Xuan Jiang}
\IEEEauthorblockA{\textit{Department of Civil and Environmental Engineering} \\
\textit{UC Berkeley}\\
Berkeley, USA \\
xuanjiang@lbl.gov}
\and
\IEEEauthorblockN{2\textsuperscript{nd} Josh Everts}
\IEEEauthorblockA{\textit{Department of Civil and Environmental Engineering} \\
\textit{UC Berkeley}\\
Berkeley, USA \\
josh\_everts@berkeley.edu}
}

\maketitle

\begin{abstract}
We used a token-wise and document-wise sentiment analysis using both unsupervised and supervised models applied to both news and user reviews dataset. And our token-wise sentiment analysis found a statistically significant difference in sentiment between the two groups (both of which were very large N), our document-wise supervised sentiment analysis found no significant difference in sentiment.
\end{abstract}

\begin{IEEEkeywords}
Electrical vehicles, unsupervised learning, supervised learning
\end{IEEEkeywords}

\section{Introduction}
Electric Vehicles (EVs) are a rapidly growing component of the automotive industry and are projected to have over 30 percent of the overall United States light duty vehicle market by 2030 \cite{b1}. Furthermore, the US and other countries have bet big on Battery Electric Vehicles (BEVs), allotting funding for charging infrastructure, subsidies and tax credits and setting deadlines to phase out combustion engine vehicles. Correspondingly, the stock price of EV companies like Tesla have recently far exceeded those of traditional auto manufacturers, helping to illustrate the bullish outlook many consumers and investors have toward EVs in general. Despite this, there remain concerns among both consumers and experts about various aspects of electric cars, and despite the excitement surrounding them, EV adoption rates hovered around 1.8\% in 2020 (energy.gov, 2021). Some oft-heard concerns include battery fires, range-anxiety, concerns about lithium and mineral mining, charging infrastructure, and indirectly-related topics such as self-driving software. We want to probe where some of these negative vs. positive sentiments arise from, and we specifically hypothesize that a difference in sentiment and topic choice can be seen between News Articles and EV consumer reviews. 

To help answer this hypothesis we will use a token-wise and document-wise sentiment analysis using both unsupervised and supervised models applied to each data set. We will then compare our sentiment values for a statistically significant difference. We will also use topic-modelling to examine what, if any, differences in topic and topic sentiment exist between news and consumer reviews. Finally, we will perform an exploratory visualization of sentiment and topics news media and consumers frequently discuss.
\section{Related Work}
While the field of NLP-based EV research is relatively young, there has been some attempt to use topic-analysis/topic-modelling and sentiment analysis/classification of user reviews. Ha et al. \cite{b2} have performed a user-experience-focused topic analysis related to EVs using language transformer models and supervised topic classification. However, whereas Ha et al. \cite{b2} focused mainly on consumer reviews of EV charging stations, we will instead focus more generally on a comparison between consumer and media sentiment towards EVs. Ha et al. found that frequent topics included charger functionality, range-anxiety, charging cost, and dealership experiences \cite{b2}. Given this, we believe that topic-modeling can be an effective way of understanding consumer data.

Another paper which focused on user-reviews of charging station experiences found by Asensio et al. \cite{b3} used a word2vec-based CNN classifier to classify sentiment. They found that positive/negative sentiment in the user reviews is roughly 50/50. Similarly to the previous work of Ha et. al., we will attempt to extend this exploration to EV-consumer reviews more broadly. 

Song et. al. \cite{b2} formulated a survey analyzing people's perceptions to Automated Vehicles (AVs). They found differences in perception and sentiment towards AV-transit systems depending on people's familiarity with technology in general. While this study is not directly related to EVs, there are many shared similarities between EVs and AVs: namely they both offer a sea-change in vehicle consumer experience. Our  project could potentially help extend and offer a broader comparison of consumer sentiment towards electric-AVs. For example, systems like 
Autopilot are directly with EV companies like Tesla.

While we have decided to focus on news and consumer-reviews for now, others have focused on social media such as Twitter. Suresha and Tawari analyzed 45000 tweets using topic models and Valence Aware Dictionary tools. They found that, on Twitter, "Tesla" was one of the top hashtags associated with EVs \cite{b4}. In related work, a Twitter-based sentiment analysis using TF-IDF scores and a Naive Bayes Classifier by Bhatnagar and Choubey found that the hashtag "\#Tesla" had a more positive sentiment than other manufacturers \cite{b5}. In light of this high association of Tesla with EVs in general, we have decided to opt to compare our news data with consumer data scraped from a Tesla forum, as it was both a large dataset and may still broadly reflect many of the topics and opinions held by those on social media as Suresha and Tawari found. 

In perhaps the most similar work to our current project, Carpenter (2015) \cite{b6} used data scraped from user discussion forums surrounding EVs to determine overarching sentiments and topics of interest. From their data, Carpenter found that "range anxiety" and "price" were two of the most common barriers to EV adoption \cite{b6}. This specific result gives us a comparison target for our own analysis and it will be interesting to see if these topics are still a concern 6 years later in 2021. Furthermore, Carpenter uses a term-frequency and regex based classification of sentiment; we attempt newer word-embedding and deep-learner based methods to see if we can generate an improvement.  

Because NLP research related to EVs is still relatively new, the overall body of work is quite small. However, from the above literature we can see that a sentiment analysis combining both user reviews and news media sentiment could provide a useful extension of previous work, as so far most literature has focused on consumer and charging reviews alone and/or yields an opportunity to apply newer supervised language learning models such as BERT.

\section{Data}
\subsection{Data Sources}
This project utilizes existing News article data taken from several business-news sites including WSJ, Reuters, CNBC, and Bloomberg. All articles scraped by Jeet were from January to May 2018 \cite{b7}. This places the time of the news data within the range of the consumer comments. We had found previously that using news headlines alone did not provide enough data, and the frequency of headlines containing EV-related information was very low ($<$0.1\%). By using business news we were able to extract more information about EVs ($\sim$1700 articles). The articles were filtered out from the main news data set if they mentioned the bi-grams "electric vehicle(s)" or "electric car(s)." 

Consumer review and opinion data was collected from insideEVs.com , which hosts a forum for EV-related discussions. We wanted a forum that had a large number of posts and featured active discussion about EV models and ownership specifically. We ended up selecting the Tesla Model 3 owners forum \cite{insideEVs} which was the largest (the Tesla Model 3 is the best-selling EV).Comments were then scraped from the forum with a Selenium script. The raw number of comments was around 1800 and they range in date from 2017 to 2021. 

\subsection{Filtering and Cleaning}
One issue with scraping forum data arises from the nested structure of many replies, as posters may copy and paste from others, or use a built-in reply feature which creates a copy of the text they respond to. It was necessary to remove this extra text from our data which we attempted with Smith-Waterman sequence alignment. We compared the text in each comment to the original poster's (OPs) text in each comment thread. If there was a sequence match, we took the largest sequence match and removed it from the comment. We then repeated this process for every comment's preceding comment. For computational reasons it was not possible to compare every comment with every other comment, though this would be optimal. Furthermore, there is a risk of accidentally 'double-counting' text alignments between comments, say with the OP, and then with some other reply. With more time, we would have liked to develop a more sophisticated algorithm to clean this data; as it is, a fair portion of the nested structure is removed.

Finally, after cleaning and removing direct comment duplicates from both data sets (likely a scraping artifact), we end up with around 1690 news articles and around 1630 comments each. Before running each document through any sentiment analysis or topic model it is first tokenized and \texttt{nltk} english stop-words are removed along with some custom stop-words which include partiular commenter-names and non-word tokens.

\subsection{Generating Training Data}

For our supervised approach, we first labelled 1000 combined news and text documents with a sentiment of $-1$ for a negative sentiment towards EVs or some aspect of EVs, $0$ for a neutral sentiment towards EVs, and $1$ for a positive sentiment towards EVs or some aspect of EVs. Then, our chosen model, TFBertForSequenceClassification \cite{b9}, is trained on this data with an 80/20 training/validation split. For the evaluation, we generated a set of "Gold Labels" which only included (document,label) pairs where both of us agreed on the label.

We opted to remove 'neutral' ($0$) sentiments from our data after labelling because most often we had used these sentiment labels when referring to data that had little relation to EVs in general. This left us with a simple binary positive/negative classification. Below is our table indicating the size (N) and \% positive for each of our classes after the data has been cleaned with neutral sentiment removed. For our un-binarized gold data we found a Cohen's $\kappa = 0.78$ indicating a substantial agreement. For our previous test-labeling, however, we had observed a $\kappa = 0.52$ only indicating moderate agreement. This may indicate that our kappa value is overall quite variable and indicates that labeling even a simple positive, neutral, or negative sentiment is not straightforward. As seen in the table below, the \% positive is generally around 50, though lower in the training data. 

\begin{table}[!ht]
    \centering
    \caption{Data proportions}
    \begin{tabular}{|l|l|l|l|}
    \hline
        ~ & Training & Validation & gold \\ \hline
        N & 529 & 139 & 60 \\ \hline
        \%+ & 50.2 & 52.5 & 43 \\ \hline
    \end{tabular}
\end{table}

\section{Methods}
We will use two main approaches in our comparison of news and consumer data: sentiment analysis and topic-modeling.
\subsection{Sentiment Analysis}
For the sentiment analysis we have opted to use both an unsupervised and supervised approach. For the unsupervised approach we will use the SemAxis \cite{b1}technique which compares the similarity of a word embedding to a sequence of user-defined positive and negative embeddings. In this way a custom 'sentiment' based off similarity to the specified sentiment vectors is generated. For the supervised approach we label our own news and review data and then train a transformer-based model, in this case TFBERT \cite{b9} on these labels. Then, the sentiment classifier is run over the news and consumer data to generate an average sentiment for each. In the table below we compare these values to the expert-annotated scores from \cite{b2}.

\begin{table}[!ht]
\caption{TFBERT Testing}
    \centering
    \begin{tabular}{|l|l|l|l|}
    \hline
        ~ & majority & gold & literature\\ \hline
        Accuracy & 57\% & 65\% &89\%\\ \hline
        F1 & NA & 0.79 & 0.82\\ \hline
        
    \end{tabular}
\end{table}
After we combined our 1600 news and 1600 comments and split them into 80 percent of train data and 20 percent of test data, we got 57 percent accuracy. And when we applied the gold data we got 65 percent accuracy and 0.79 F1 score.

For the SemAxis approach we will use two embeddings: untrained GloVe embeddings and word2vec embeddings each trained on the news and consumer data, respectively. Then, we perform a token-wise mean of all the embeddings in the news and consumer data to get an average sentiment for each corpus. We can then compare these values to each other using a Student's t-test to see if they are statistically different from one another. 

\subsection{Topic Modeling}
For our Topic Modelling we will be making use of the gensim topic-model library, specifically the gensim Latent Dirichlet Allocator (LDA) model. We have opted to use K=20 for this model, meaning it will generate 20 probable topics for each of our corpora. Then, we perform a word2vec embedding-based SemAxis analysis on each set of topic model tokens to get an idea of the sentiment distribution over each topic. 

\section{Analysis}

\subsection{Comparing News and Consumers Categorically}

We begin with the significance testing of our token-wise sentiments for our unsupervised SemAxis model and the document-wise sentiments for our supervised TFBERT model.
\begin{table}[!ht]
\caption{Topic Seperations}
\scalebox{0.2}{
\centering
\begin{tabular}{|c|c|c|c|c|c|}
\hline
    Corpus & ~ & Main Topic 1 & Main Topic 2 & Main Topic 3 & Main Topic 4 \\ \hline
    News & ~ & mining/environment & stocks and production & self-driving & trade with China \\ \hline
    ~ & \% total News & 35\% & 25\% & 20\% & 15\% \\ \hline
    ~ & Example & lithium china demand supply tonnes project government mining copper production chile companies & tesla model 3 production musk quarter billion shares week analysts per cash & tesla nickel crash autopilot driver system fire police model battery safety may & cars china technology one people industry world business could billion selfdriving first \\ \hline
    Comments & ~ & car features and comparisons & stocks and production & cold-weather range & environment \\ \hline
    ~ & \% total Comments & 50\% & 25\% & 20\% & 5\% \\ \hline
    ~ & Example & regen bolt driving car 3 tesla youtube never go model ’ rear control & production tesla numbers model week going 'm 3 curve 2018 get deliveries time & battery tires temperature know range v charging pack car tesla've front charge & nuclear one waste good new fuel energy years plants us tesla probably fossil \\ \hline
\end{tabular}
}
\end{table}

We can see from Table 4 that the overall sentiment values for the gloVe and word2vec embeddings are near-0, but that they are both statistically significant over the total number tokens in each corpus ($\sim$80,000 in comments and $\sim$560,000 in news). Meanwhile our tfBERT model shows no chi-squared statistical significance. Overall this suggests little to no difference between the news and consumer data.

\begin{table}[!ht]
\caption{Model Results}
\scalebox{0.8}{
    \centering
    
    \begin{tabular}{|l|l|l|l|l|l|}
    \hline
        Model & News mean & News std & Comments mean & Comments std & p $\ge$ (u1!=u2) \\ \hline
        gloVe & 0.10 & 0.094 & 0.18 & 0.123 & $\le$ 0.0001 \\ \hline
        word2vec & -0.015 & 0.109 & -0.056 & 0.141 & $\le$ 0.0001 \\ \hline
        tfSequenceBERT & -0.106 & NA & -0.118 & NA & $\chi^{2}$: 0.74 \\ \hline
    \end{tabular}
    }
\end{table}

We can see from Table 3 that even when trying to shrink around K=20 topics into 4 main topics, the main topics did not strictly align. We also see that, while the overall percentages for the "stock and production" main topic aren't that different, the topic "mining and environment" has very different percentages between the two. This is an indicator that, in general, Tesla owners are not typically discussing metal mining or general environmental impacts (positive or negative) of their cars relative to these business-news media.

\subsection{Connecting Sentiment to Geography}

One thing we were interested in exploring was what sentiments were most associated with particular cities and regions in the news. Fig. 1 displays the top 5 most common entities in our News data and their word2vec News Sentiment embedding. 

\begin{figure}[h!]
  \caption{Most Common Locations and Their News Sentiment}
  \includegraphics[scale=0.2]{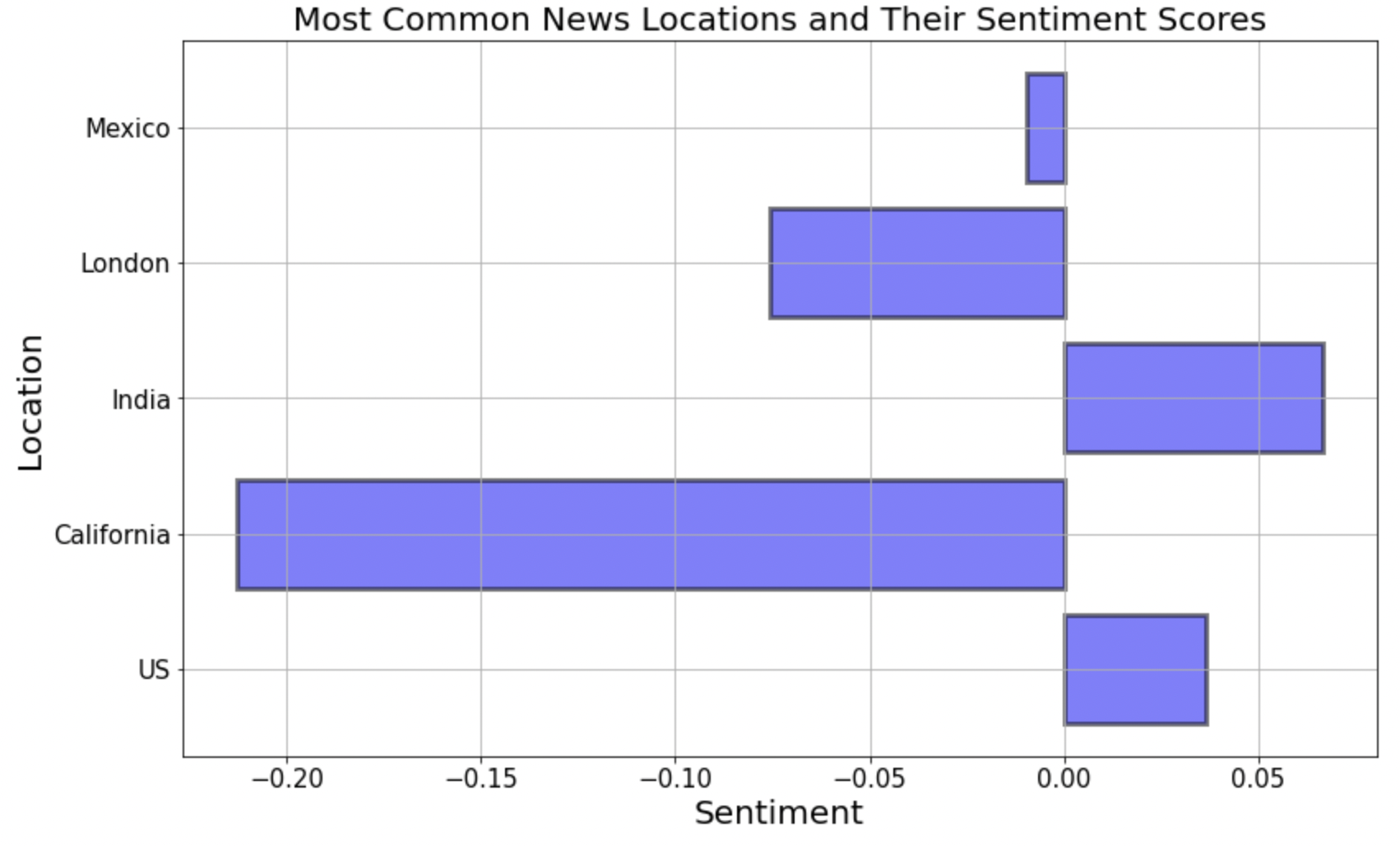}
\end{figure}

We can see from the figure that California has a very negative sentiment relatively speaking, while India has a fairly positive sentiment. We think that the negative sentiment displayed by California could reflect the fact that much of the time articles with negative sentiment surrounding Tesla stock and autopilot crashes directly mention California. Meanwhile, we think that the fairly positive sentiment surrounding India is due to the general excitement companies and investors see in what is an untapped EV market. While this analysis is interpretive, we thought it was quite striking to see such a large difference in sentiment for a geographic entity. 

\subsection{Understanding Sentiment Through Topic}

\begin{figure}[h!]
  \caption{Sentiment Distribution Across 20 Topics}
  \includegraphics[scale=0.3]{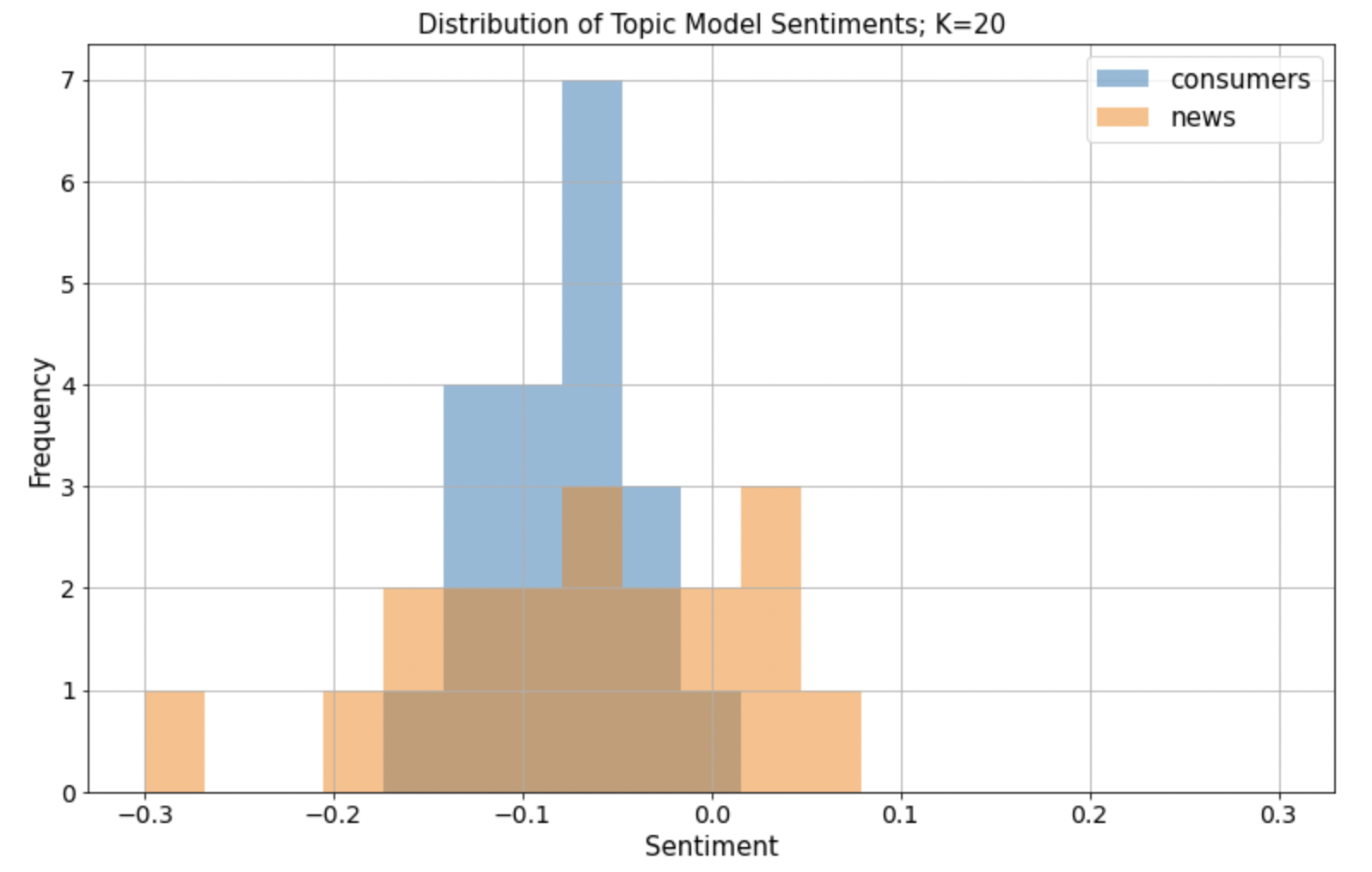}
\end{figure}

From Fig. 1 we can see that the overall sentiment distributions are visually quite different. The consumer review data has a negative sentiment, but is more tightly clustered than the news topic sentiments. Interestingly, there is one sentiment which is quite negative (around $-.30$) in the news data. This distribution suggests that the difference in sentiment between news and consumer topics is not so much a difference in \emph{mean} sentiment value, but rather sentiment spread. 

We can also try and understand consumer sentiment through a topic-model lens. Since we have already seen that a topic model can offer some insight into the sentiment distribution of users and news -- more than the sentiment scores alone do -- we can also just choose to look at sentiments directly. Below in Fig. 2 and Fig. 3 we display a time-oriented look at the frequency of a particular topic over time. 

\begin{figure}[h!]
  \caption{Vehicle Price Topic Over Time}
  \includegraphics[scale=0.3]{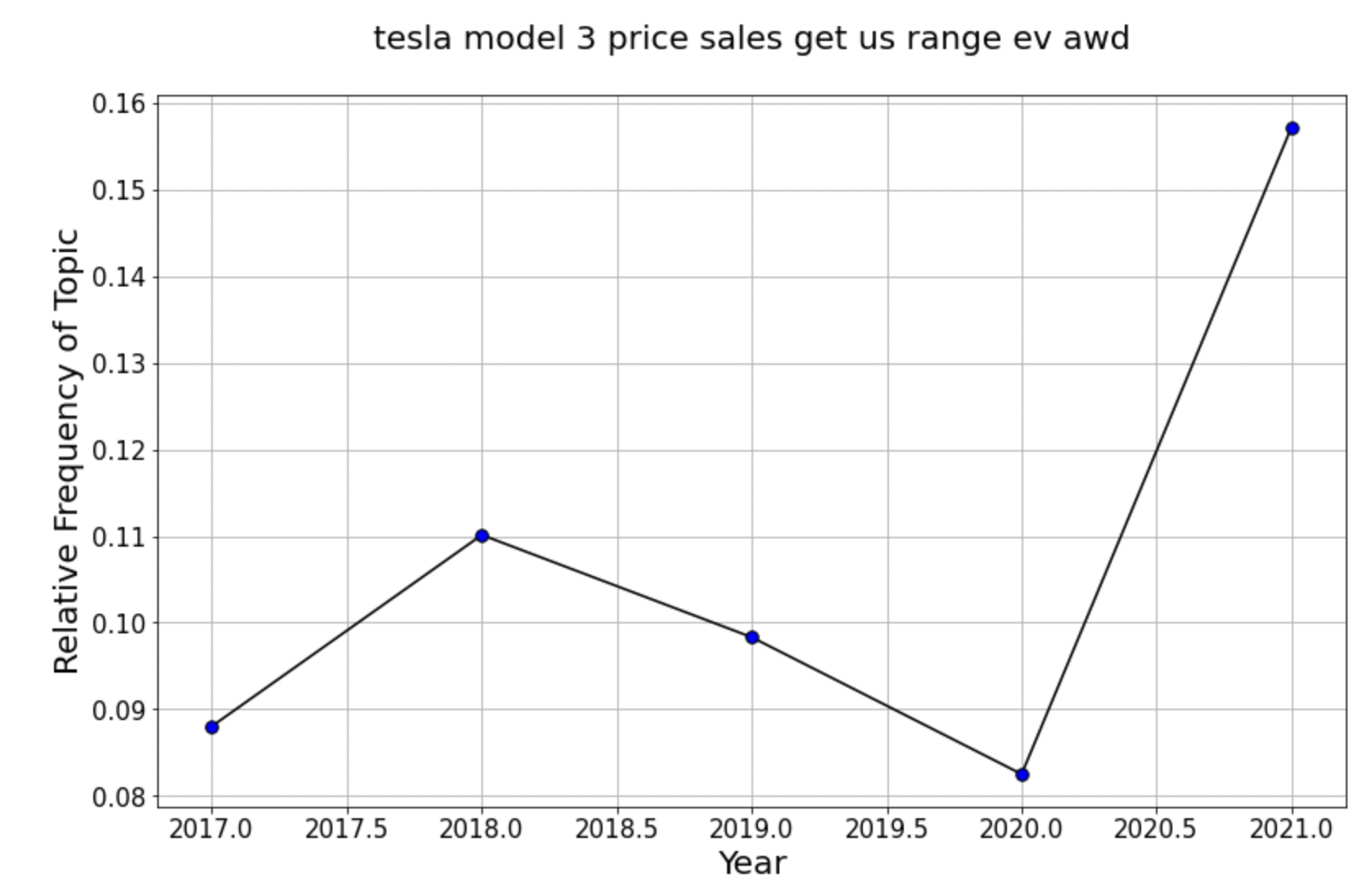}
\end{figure}
\begin{figure}[h!]
  \caption{Tesla Competition Over Time}
  \includegraphics[scale=0.3]{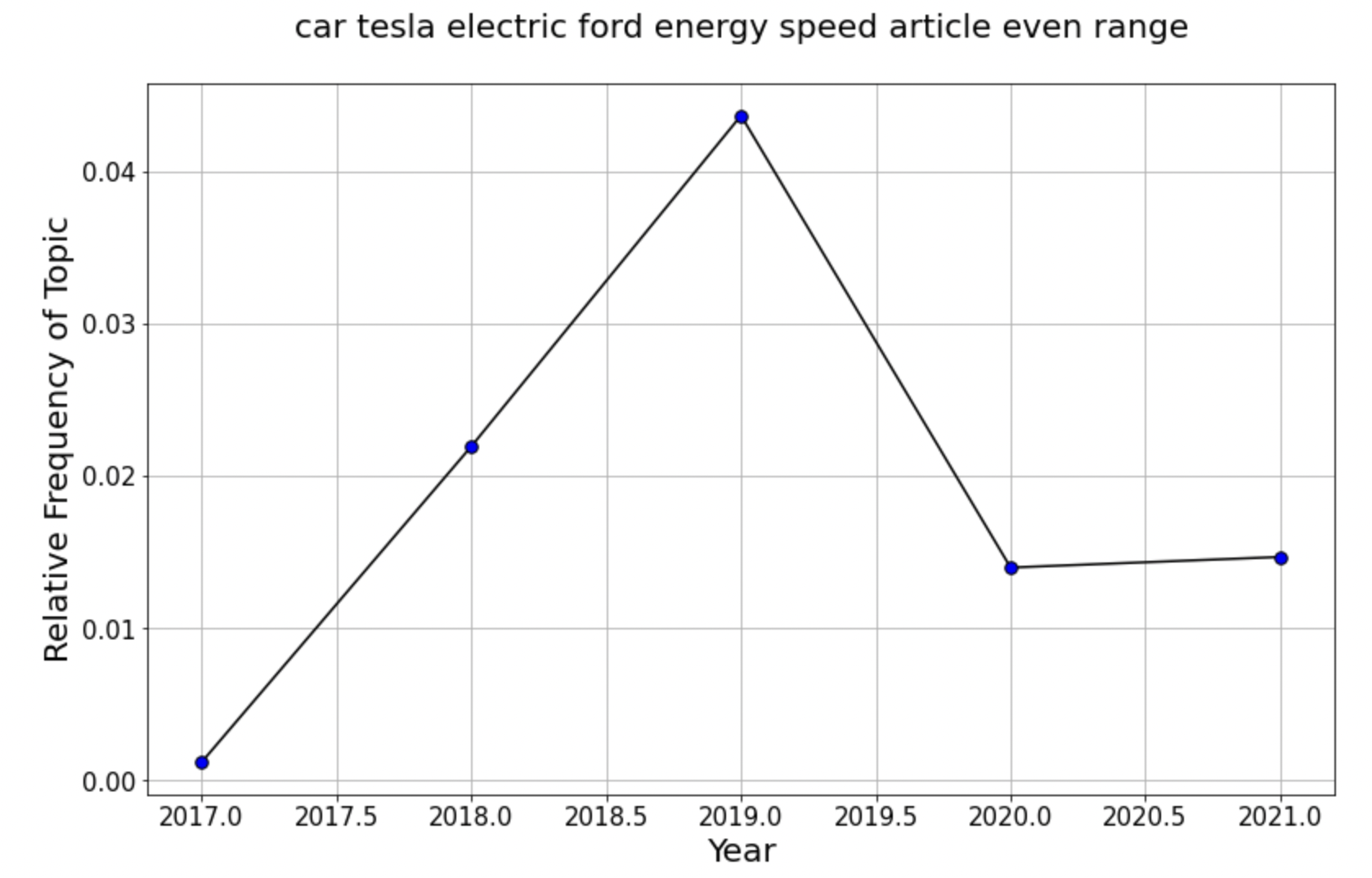}
\end{figure}

Looking at Fig. 2 we can see that from 2017 to 2021, this topic concerning vehicle prices and sales has varied. Firstly, in 2018 it receives a spike when the Tesla Model 3 was first coming to mass-market and availability and pricing information were sometimes scarce. Then in 2021 as both the prices of cars and of EVs have increased the topic has rebounded in frequency. 

Likewise, in Fig. 3 we can see that 2019 is the peak year for this topic, the year where Ford unveiled it's Mustang Mach-E EV. During this time people were frequently discussing potential upcoming competition with Tesla an debating what vehicle was better. This is a very specific topic, but one which the topic model was able to find without much trouble.

This topic-modelling provides a more direct window into consumer sentiment over time, especially concerning specific emotions like excitement over a new Tesla competitor. Topic models were likely chosen by much of the existing EV NLP literature precisely for this reason, and if we were to invest further time into the project, Transformer-based topic-modeling would be an interesting next step. 

\section{Conclusions}
\subsection{Does a sentiment difference really exist in our data?}
One of the main goals of our project was to determine if there really was a difference in sentiment between news articles and consumer/owner reviews and comments. In order to test this hypothesis we attempted both an unsupervised and supervised approach using a business-news data set and scraped Tesla Model 3 forum comments. While our token-wise sentiment analysis found a statistically significant difference in sentiment between the two groups (both of which were very large N) our document-wise supervised sentiment analysis found no significant difference in sentiment. Furthermore, the practical significance of the token-wise sentiments was very small and highly dependent on choice of semantic vector. Therefore, despite limitations of our data and model, we can say that there is likely no overall difference in sentiment between business-news and comments from active, engaged, Tesla Model 3 owners. 
\subsection{Limitations of Our Data and Model}
Due to the relatively small starting size of our combined raw data set at around 3600 total documents, filtering and further sub setting of the data ended up potentially affecting our model more than we would have liked. This is best illustrated by our small gold-test class which had only 60 documents. This likely affected the overall model accuracy, as it is possible that due to random chance alone our gold data was not as representative of our training and development data. However, we believe that this is probably not the largest limitation on our results.

For our unsupervised SemAxis analysis, the largest limitation is the subjectivity of defining a semantic axis in the first place. It is likely that other people wouldn't choose or find the same words to use as positive or negative sentiment vectors, thereby limiting the overall generalizability of our results. In our supervised analysis we also found that accurately classifying sentiment towards or against EVs was very difficult; the tendency was to extrapolate what a particular choice of wording or topic might have meant about that author's sentiment towards EVs regardless if that sentiment was truly easy to spot or not. This is an issue that could be resolved with more annotators and a larger data set, allowing for both training and testing on data that has multi-annotator agreement. 

Finally, perhaps the biggest challenge and hurdle in our analysis was cleaning the actual documents themselves. Early on we realized that many of the user forum comments had repeat text from other comments distributed throughout. While Sequence Alignment partially resolves this issue, the choice to automate the removal of repeat sequences has to be done carefully as we learned. It is very easy to double-count alignments from other comments and remove someone's contribution entirely from the corpus. With more time and computational power we think that we could use the same Smith-Waterman alignment even more effectively. 

Additionally, we realized that many news articles have very similar or identical text. While we were able to remove some duplicates from our data, we did not have the time or computational power to do the necessary sequence alignment. In this case we think the news data set may have mistakenly scraped some duplicate comments, and that some news organizations may have very similar automated reports (about stocks for example). In further work this repetition in the news data would be something to look for early on. 

\subsection{What can NLP reveal about EVs?}

While there are certainly limitations to our data and model, and our initial hypothesis that there would be a difference in sentiment between consumers and news media had little support, there are still some interesting takeaways from our project. 

Firstly, with small numbers of words or individual words, SemAxis can provide interesting insights into the data. In our topic-model sentiment analysis, for example, we can clearly see that the overall sentiment distribution of news topics is wider than those of tesla forum consumers. While perhaps surprising, it does make sense that consumers are more likely to discuss topics in a neutral, information-seeking way on a forum in many cases. Additionally, as consumers of EVs, Tesla Model 3 forum-goers are not likely to post about unsavory mining and labor practices or dips in Tesla stock price, as they are there to discuss and enjoy their community. We think this shows that, on the whole, news articles often try to investigate a specific viewpoint about EVs, either positive or negative, regardless if that viewpoint is discussed by or interesting to EV owners. 

Additionally, we see in our consumer review topic-models that discussions of issues like range, battery-health, and build-quality continue to be relevant on the Tesla Model 3 forum even in 2021. Many of these topics were also found analyzed by \cite{b6} in 2015. This indicates that significant progress still needs to be made on these issues and that they are still concerns and barriers to widespread EV-adoption.

\vspace{12pt}

\end{document}